\documentclass[a4paper]{article}

\usepackage{INTERSPEECH2020}

\usepackage{url}
\usepackage{hyperref}
\usepackage{graphicx}
\usepackage{amssymb,amsmath,bm}
\usepackage{algorithm2e}
\usepackage{textcomp}
\usepackage{comment}
\usepackage{multirow}
\usepackage{stfloats}

\newcommand{\tabincell}[2]{\begin{tabular}{@{}#1@{}}#2\end{tabular}}
\def\vec#1{\ensuremath{\bm{{#1}}}}

\title{AutoSpeech 2020: The Second Automated Machine Learning Challenge for Speech Classification}

\makeatletter
\def\name#1{\gdef\@name{#1\\}}
\makeatother
\name{{\em Jingsong Wang$^1$, Tom Ko$^{2*}$, Zhen Xu$^1$, Xiawei Guo$^1$, Souxiang Liu$^1$, Wei-Wei Tu$^{1,3}$, Lei Xie$^4$} \thanks{* corresponding author}} 

\address{$^{1}$4Paradigm Inc., Beijing, China  \\
  $^2$Department of Computer Science and Engineering\\ Southern University of Science and Technology, Shenzhen, China \\ 
  $^3$ChaLearn, USA \\
  $^4$ASLP@NPU, Northwestern Polytechnical University, Xian, China
}
\email{wangjingsong@4paradigm.com, tomkocse@gmail.com}

\begin{document}

\maketitle

\begin{abstract}
The AutoSpeech challenge calls for automated machine learning (AutoML) solutions to automate the process of applying machine learning to speech processing tasks. 
These tasks, which cover a large variety of domains, will be shown to the automated system in a random order.
Each time when the tasks are switched, the information of the new task will be hinted with its corresponding training set.
Thus, every submitted solution should contain an adaptation routine which adapts the system to the new task.
Compared to the first edition, the 2020 edition includes advances of 1) more speech tasks, 2) noisier data in each task, 3) a modified evaluation metric.
This paper outlines the challenge and describe the competition protocol, datasets, evaluation metric, starting kit, and baseline systems.


\end{abstract}

\section{Introduction}
\label{sect:intro}
\vspace{2mm}
In the past few decades, machine learning, especially deep learning, has achieved remarkable breakthroughs in a wide range of speech tasks, e.g., speech recognition \cite{povey2016purely,chan2016listen}, speaker verification \cite{garcia2019speaker,zhu2019mixup,ko2020prototypical}, language identification \cite{muralikrishna2019spoken,grisard2019spoken} and emotion classification \cite{albornoz2014spoken,morrison2007ensemble}. 
Each speech task has its own specific techniques in achieving the state-of-the-art results \cite{garcia2019speaker,muralikrishna2019spoken,albornoz2014spoken,weninger2019deep,dobry2011supervector,fu2010survey}, which require efforts of a large number of experts.
Thus, it is very difficult to switch between different speech tasks without human efforts.
In fact, a lot of speech tasks use similar techniques in feature extraction, model selection, optimization, etc.
Therefore, if there is an automated way to encapsulate different speech tasks in the same framework and enhance the sharing of overlapped techniques, the cost for non-experts in solving speech problems will be greatly reduced.

Automated machine learning (AutoML) aims at automating the process of applying machine learning to real-life problems \cite{yao2018taking}.
Till now, it has been successfully applied to many important problems, e.g., neural architecture search \cite{zoph2016neural,liu2018progressive}, automated model selection \cite{feurer2015efficient,kotthoff2017auto} and feature engineering \cite{katz2016explorekit,kanter2015deep}.
All these successful examples serve as a ground for the possibility of applying AutoML to the field of speech.

To foster research in the area of AutoML, a series of AutoDL competitions\footnote{https://autodl.chalearn.org}, e.g., Automated natural language processing\footnote{https://www.4paradigm.com/competition/autoNLP2019} (AutoNLP) and Automated computer vision\footnote{https://autodl.lri.fr/competitions/3} (AutoCV2), have been organized by 4Paradigm, Inc. and ChaLearn (sponsored by Google). 
These competitions, proposed to explore automatic pipelines to train an effective DL model given a specific task requirement, have drawn a lot of attention from both academic researchers and industrial practitioners. 

Autospeech 2020 Challenge\footnote{https://www.4paradigm.com/competition/autospeech2020} is the second in a series of automated speech challenges, which applies AutoML to the tasks in speech processing. 
Unlike many challenges \cite{ryant2019second,todisco2019asvspoof}, we require code submission instead of prediction submission. 
Participants' codes will be automatically run on multiple datasets on competition servers with the same hardwares (CPU, GPU, RAM, etc.) in order to have fair comparisons.
Participants should strike a balance between the effectiveness and the efficiency of their solutions and the codes should not be hard to deloy.
All the datasets are split into training and testing parts.
The private datasets, including their training and testing parts, are unseen by the participants.
Instead of evaluating with datasets from a single application, our evaluation is done on datasets from different applications in this competition.
Top ranked solutions under this setting should have good generalization ability.

We use Anytime Learning metric, which considers the whole learning trajectory, instead of the traditional metric, which focuses on the converged performance only.
In our challenge, we use Area under Learning Curve (ALC), which is an integral of the learning curve (whose points are balanced ACC of predictions at different timestamp) \cite{liu2019autocv}.
From our experience, ALC is more suitable for a challenge and is closer to real application requirement.

The first AutoSpeech Challenge (AutoSpeech 2019\footnote{https://www.4paradigm.com/competition/autospeech2019}) was held in ACML 2019 and was part of the AutoDL challenge in NeurIPS 2019. 
It attracted 33 teams. 
The top ranked teams adopted different automatic learning strategies, including 
model pre-training and multi-model ensembling.
With feedback from the AutoSpeech2019 challenge, we modified a couple of settings in order to make the AutoSpeech2020 more interesting and challenging. 
First, we remove a few simple tasks and replace them with more difficult tasks, in which there are fewer samples or more categories. 
Then, we increase the time budget (which was originally 20 minutes) in order to encourage more complex methods.
Last but no least, we change the evaluation index from AUC to balanced ACC at each timestamp of the learning curve. 
It is because balanced ACC can better indicate the classification ability of the models when they are over fitted to a few categories.
With the above changes, we believe AutoSpeech2020 will be more enjoyable, challenging and real-life oriented.

The paper is organized as follows: Section \ref{sec:proto} describes the design of the competition, including competition protocol, datasets, metrics and starting kit. Section \ref{sec:baseline} describes the baseline we use and results of the  experiments. Section \ref{sec:conclusion} presents the conclusions.

\section{Competition Design}
\label{sec:proto}
\vspace{4mm}

\subsection{Competition protocol}
\label{subsec:protocol}

AutoSpeech 2020 Challenge adopts the similar competition protocol of the AutoDL Challenge \cite{liu2018autodl}, in which there are three phases: Feedback, Check, and Final. 
In the Feedback Phase, the participants are provided with five practice datasets which can be downloaded, so that they can develop their solutions offline. 
Then, the codes will be uploaded to the platform and participants will receive immediate feedback on the performance of their methods upon another five feedback datasets. 
Note that participants cannot see the examples and the labels of the feedback datasets.
After the Feedback Phase terminates, there will be the Check Phase, in which participants are allowed to submit their codes only once to run on the private datasets in order to debug. 
Participants are not able to read detailed logs but reported errors. 
In the Final Phase, participants' solutions will be evaluated on five test datasets.
The ranking in the Final Phase will determine the winners.

Submitted codes are trained and tested automatically, without any human intervention. 
In the Feedback (resp. Final) Phase, they are run on all five feedback (resp. final) datasets in parallel on separate compute workers, each one with its own time budget. 
The identities of the datasets used for testing on the platform are concealed. 
The data are provided in a raw form (no feature extraction) to encourage researchers to use Deep Learning methods performing automatic feature learning (although this is NOT a requirement). 
All problems are multi-label classification problems. 

\begin{figure}[htb] 
\centering 
\includegraphics[width=0.45\textwidth]{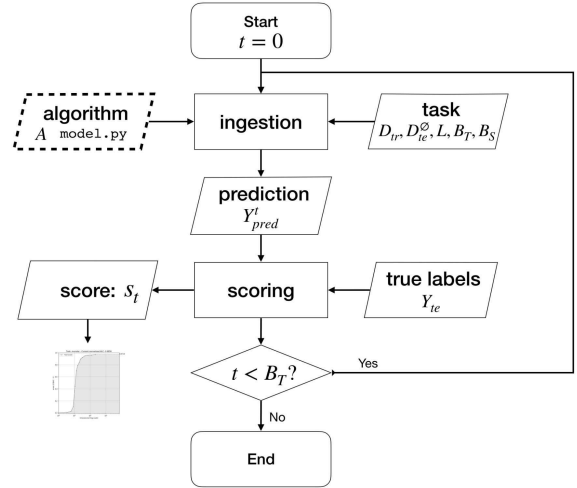} 
\caption{
\textbf{AutoSpeech Challenge's evaluation process} for one task defined by the 5-tuple: $D_{tr}, D^{\emptyset}_{te}, L, B_T, B_S$. Participants need to submit a strategy implemented by Python scripts which must contain a file named "$\mathsf{model.py}$". 
forgiving In this file, the two functions named $\mathsf{train}$ and $\mathsf{test}$ implement the logic of AutoSpeech algorithm. These two functions are called by the ingestion program (defined in $\mathsf{ingestion.py}$) orderly, to train on $\textit{D}_{tr}$ and produce a prediction $Y^t_{pred}$ on $D^{\emptyset}_{te}$ at the timestamp $t$ respectively. The prediction $Y^t_{pred}$ is then compared to true labels $Y_{te}$ in scoring program (defined by $\mathsf{score.py}$ and produces a score $s_t$. This ingestion/scoring loop is controlled by time budget $B_T$ and a flag in $\mathsf{model.py}$, to determine whether the whole program ends. At any time,the score sequence ${s_t}_0$, ${s_t}_1$, ... is visualized as a learning curve and the area under learning curve is used as the evaluation for this task. Note that only the run time of ingestion program is accumulated, but the time of scoring is not counted for running in parallel.  }
\label{Fig.evaluation-process} 
\end{figure}

For a single task, the evaluation process is shown in Figure \ref{Fig.evaluation-process}, which is the same as the AutoCV Challenge \cite{liu2019autocv}. The task, which has the same definition, is defined by 5-tuple:

$$\mathcal{T} = (D_{tr}, D^{\emptyset}_{te}, L, B_T, B_S)$$

where $\textit{D}_{tr}$ and $\textit{D}_{te}$ are separated from a dataset $D = {(x_i,y_i)}^n_{i=1}$, $D^{\emptyset}_{te}$ and $Y_{te}$ are examples and labels in test set, $L:\mathcal{Y} \times \mathcal{Y} \rightarrow \mathbb{R} $ is a loss function measuring the losses $L(y',y)$ of the predictions $y'$ with respect to the true labels $y$, and $B_T$ and $B_S$ are time and space budget restrictions respectively. $B_T$ of each dataset is limited to a maximum of 30 minutes (with extra 20 minutes of initialization). The above definition is applicable to many kinds of AutoML Challenges, and this time, all the tasks focus on speech.

\subsection{Datasets}
\label{subsec:config}
As mentioned above, there are 3 types of datasets: practice datasets, feedback datasets, and private datasets in this challenge, and each of them contains 5 datasets.
Five practice datasets, which can be downloaded, are provided for the participants to develop their AutoSpeech solutions offline. 
Besides that, another five feedback datasets are provided for participants to evaluate the public leaderboard scores of their AutoSpeech solutions. 
Afterwards, their solutions will be evaluated with five private datasets without human intervention.

Each provided dataset is from one speech classification domain, including Speaker Identification, Emotion Classification, Language Recognition, etc. 
Each dataset is obtained from unbalanced sampling in only one task.
In the datasets, the number of classes is greater than 2 and fewer than 500, while the number of instances varies from several to hundreds. 
All the audios are first converted to single-channel, 16-bit streams at a 16kHz sampling rate for consistency. Then they are loaded by Librosa and dumped to pickle format (A list of vectors, which contains all training or testing audios in one dataset). 
The datasets containing both long audios and short audios are without padding. Table \ref{tbl:practice-dataset} shows the summary of the practice datasets, which are from 5 task domains: Speaker Recognition\cite{panayotov2015librispeech}, Emotion Recognition\cite{burkhardt2005database}, Accent Identification \cite{weinberger2015speech}, Music Genre Classification \cite{ellis2007classifying} and Spoken Language Identification \cite{park2019css10}.

\begin{table*}[htb]
  \centering
  \setlength{\abovecaptionskip}{5pt}
  \caption {\it \textbf{Autospeech practice datasets summary.} We provide 15 datasets and 5 of them can be downloaded for local debugging. Each dataset is sampled from a specific task domain, and is ensured that the number of samples in each category of the training set and the testing set is relatively balanced. Besides, the last column is the total size of each dataset.}
   \begin{tabular}{c  c  c  c  c  c  c}	\hline
   \hline
	  \textbf{No.} & \textbf{Dataset Source} & \textbf{Task} & \textbf{Class Number} & \textbf{\tabincell{c}{Training Set \\ Number}} & \textbf{\tabincell{c}{Testing Set \\ Number}} & \textbf{Size} \\
      \hline
      \hline
      01 & Librispeech & speaker & 330 & 1650 & 3300 & 306M \\
      \hline
      02 & Berlin & emotion & 7 & 346 & 162 & 87M \\
   	  \hline
      03 & Speech Accent Archive & accent & 11 & 164 & 308 & 824M \\
   	  \hline
   	  04 & artist20 & music genre & 20 & 343 & 739 & 2.0G \\
   	  \hline
   	  05 & CSS10 & language & 10 & 132 & 151 & 35M \\
   	  \hline
   	  \hline
   \end{tabular}
  \label{tbl:practice-dataset} 
\end{table*}

\subsection{Metrics}
\label{subsec:config}

\begin{figure}[htb] 
\centering 
\includegraphics[width=0.5\textwidth]{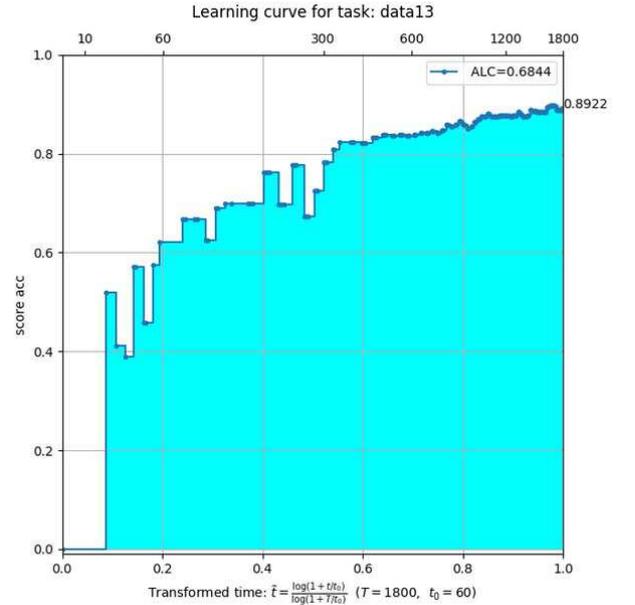} 
\caption{\textbf{Example of learning curve} shows the performance as a function of time. The strategies submitted by participants produce a sequence of predictions over time until the time limit is attained (\ref{subsec:protocol}). The curve is a piecewise function and the area under it is used as the evaluation index of the current task. As shown in the figure, the time interval between two predictions can be very short, and a new round prediction can be carried out with only minor adjustments. Of course, a large time interval or no prediction in progress might be possible, when it is over-fitting or no better prediction is worth producing.} 
\label{Fig.learning-curve} 
\end{figure}

AutoSpeech challenge encourages any-time learning by scoring participants with the Area under the Learning Curve (ALC)(Figure \ref{Fig.learning-curve}). 
In the specified time of each task, participants can carry out incremental training for multiple times, and verify the current model effect, which calculate the “performance” of the learning curve.
So over time, the points on the curve will gradually become better, and tend to be stable.
Achieving better results in a shorter time helps obtain a larger area under the curve, which means a better score. 
More precisely, for each prediction made at a timestamp when the participant’s strategy decides to conduct a test, we calculate the balanced accuracy which is used to draw a point of curve. 
Then, the learning curve is drawn as follows:

\begin{itemize}
    \item at each timestamp t, we compute s(t), the balanced accuracy of the most recent prediction. In this way, s(t) is a step function w.r.t time t;
    \item in order to normalize time to the [0, 1] interval, we perform a time transformation by
    
    $$\tilde{t}(t) = \frac{\log (1 + t / t_0)}{\log (1 + T / t_0)}$$
    
    where T is the time budget and t0 is a reference time amount (of default value 60 seconds).
    \item then compute the area under learning curve using the formula
    
    \begin{equation*}
    \begin{aligned}
    ALC &= \int_0^1 s(t) d\tilde{t}(t) \\
    &= \int_0^T s(t) \tilde{t}'(t) dt \\
    &= \frac{1}{\log (1 + T/t_0)} \int_0^T \frac{s(t)}{ t + t_0} dt \\
    \end{aligned} 
    \end{equation*}
    
    we see that s(t) is weighted by 1/(t + t0)), giving a greater importance to prediction made at the beginning of the learning curve.
\end{itemize}

After we compute the ALC for all datasets, the overall ranking is used as the final score for evaluation and will be used in the leaderboard. 
It is computed by averaging the ranks (among all participants) of ALC obtained on the datasets.

\subsection{Starting kit}
\label{subsec:starting-kit}

We provide the participants with a starting kit, which contains toy sample data, baseline submission code, and the ingestion and scoring code that has the similar call logic with the online challenge platform, for participants. Participants can create their own code submission by just modifying the file "model.py" or adding other dependency code files, pre-train models, and then upload the zip-package of the submission folder. It is very convenient to test and debug strategy locally with the same handing programs and Docker image of the Challenge platform, and evaluate strategy progress by experimenting with practice datasets. Starting kit can be run in both CPU and GPU environment, but the version of cuda cannot be lower than 10 if GPU is used. Participants can check the python version and install python packages in the docker of starting kit.

\section{Baseline and Experiments}
\label{sec:baseline}
\vspace{4mm}

\subsection{Baseline method}
\label{subsec:baseline-method}

AutoSpeech 2020 adopts the method of the champion in the first AutoSpeech Challenge, a team (PASA\_NJU) from Nanjing University PASA Lab for Big Data, as the baseline method. 
Their solution focuses on model selection and result ensemble.
As this is a challenge with code submission, we provide the baseline code in the starting kit (Sec.\ref{subsec:starting-kit}), so participants can experiment and improve on the basis of the baseline. 
In order to make the problem more rigorous and challenging, we make the improvements mentioned in Section \ref{sect:intro}.
We hope to get more innovative and excellent auto strategies for AutoSpeech problem.


The baseline code contains complex logic to control data sampling, feature engineering, model selection, training process, and result ensemble. The main strategies are as follows. 
First of all, the whole training process is divided into many rounds, that is, calling the interface of train and test functions several times in the evaluation process (Sec.\ref{subsec:protocol}). 
There is a model library, including Logistic Regression (LR), Convolutional Neural Networks (CNN), Recurrent neural networks (RNN), etc., where the strategy selects a sub-set of models according to certain rules for training. 
The same model structure may be trained many times, and many rounds in each training process.
After each round of training, whether the current model is over-fitting is judged according to the effect of validation set.
Meanwhile, splitting data, sampling, and feature extraction are carried out several times when doing incremental training.
As long as the performance of the validation set meets the requirement, the prediction of test dataset will be added onto ensemble set, waiting for the final screening.
In addition, the ensemble set is also dynamically updated to save the best batch of prediction results.
Besides, as the evaluation index is the area under the curve, it will train the models with shorter time first and generate the results in time, and then gradually improve the overall performance.

\subsection{Experiments on practice and feedback datasets}
\label{subsec:baseline-experiments}

We run above baseline method on the formatted datasets. All these experiments are carried out on Google Cloud virtual machine instances under Ubuntu 18.04, with one single GPU (Nvidia Tesla P100) running CUDA 10 with drivers cuDNN 7.5, 100 GB  Disk and 26 GB Memory.The time budget is fixed to 30 minutes for all tasks.
The results on practice and feedback datasets are presented in Table \ref{tbl:result}. 

As shown in Table \ref{tbl:result}, the performance of each task is quite different, because of the differences of different tasks, including domains of tasks, number of samples, number of classes, duration of each audio, etc. Though the baseline
can perform well on relatively easier datasets, there are a lot of room for improvement with more realistic tasks.
Therefore the participants are expected to submit more flexible and robust strategies.

\begin{table}[htb]
  \centering
  \setlength{\abovecaptionskip}{10pt}
  \caption {\it \textbf{Baseline Result} on practice and feedback datasets used in AutoSpeech 2020. Performance of 10 dataset from 6 domain of task, are balanced ACC and Area under Learning Curve (ALC), with a time limit of 30 minutes. }
   \begin{tabular}{c | c | c | c | c}	\hline
   \hline
	   & \textbf{Dataset} & \textbf{\tabincell{c}{Task \\ Domain}} & \textbf{\tabincell{c}{balanced \\ ACC}} & \textbf{ALC} \\
      \hline
      \hline
      \multirow{5}{*}{practice} & 01 & speaker & 0.1852 & 0.1524\\
         & 02 & emotion & 0.8642 & 0.8371 \\
         & 03 & accent & 0.1851 & 0.1552\\
         & 04 & genre & 0.5291 & 0.4306\\
         & 05 & language & 0.8675 & 0.857\\
   	  \hline
   	  \multirow{5}{*}{feedback} & 11 & speaker & 0.1670 & 0.1220\\
         & 12 & emotion & 0.8484 & 0.7917 \\
         & 13 & age & 0.4401 & 0.4196 \\
         & 14 & genre & 0.5085 & 0.4206 \\
         & 15 & language & 0.112 & 0.1084\\
   	  \hline
   	  \hline
   \end{tabular}
  \label{tbl:result} 
\end{table}

\section{Conclusions}
\label{sec:conclusion}
\vspace{4mm}

AutoSpeech 2020 focuses on Automated Machine Learning for speech classification tasks. We upgrade the challenge this year, according to the feedback from the first competition, including more speech tasks, noisier data and changing the evaluation metric.
In this paper, we outline the challenge and describe the competition protocol, datasets, metrics, starting kit and baseline. 
The baseline is the champion code of the first AutoSpeech Challenge. 
We have tested the baseline methods and shown the experiments on practice and feedback datasets. 

From the results of the first challenge and the performence of the baseline this year, it is still challenging in this problem. 
There is a big gap between the state-of-the-art results of each task in their own domain, and the results provided by a more general AutoSpeech strategy for all speech classification tasks. 
We need more efforts and experiments to answer this question about how to automatically extract useful features for different tasks from speech data, how to automatically discover various kinds of paralinguistic information in spoken conversations, and how to apply the technology of AutoML more sufficiently in the field of speech.
Meanwhile, because of instability, good performance is hard to achieve when various tasks are handled by a single solution.
Now the feedback phase is over and the rest is in progress. We expect participants to obtain better performance with advanced settings in the 2020 edition. 
The result will be reported at Interspeech 2020 and feedback from the community will be sought.

\section{Acknowledgements}
\label{sec:acknoledgements}
\vspace{4mm}
This project was supported in part by 4Paradigm Inc., ChaLearn and Google Zurich.
The authors would like to thank Hugo Jair Escalante, Isabelle Guyon and Qiang Yang for guidance as advisors.
The platform, automl.ai\footnote{https://www.automl.ai}, is built based on Codalab\footnote{https://competitions.codalab.org}, an web-based platform for machine learning competitions \cite{weinberger2015speech}.


\clearpage

\bibliographystyle{IEEEtran}

\bibliography{refs}

\end{document}